\pdfoutput=1

\documentclass[11pt]{article}

\usepackage{acl}

\usepackage{times}
\usepackage{latexsym}

\usepackage[T1]{fontenc}

\usepackage[utf8]{inputenc}

\usepackage{microtype}

\usepackage{graphicx}
\usepackage{amsmath}
\usepackage{amssymb}
\usepackage{multirow}
\usepackage{arydshln}
\usepackage{bm}
\usepackage{enumitem}
\usepackage{booktabs}

\title{Discourse Cohesion Evaluation for Document-Level \\ Neural Machine Translation}

\author{Xin Tan$^{1,2}$, Longyin Zhang$^{1,2}$, Guodong Zhou$^{1,2}$\thanks{~~Corresponding author} \\
  1. Institute of Artificial Intelligence, Soochow University, China \\
  2. School of Computer Science and Technology, Soochow University, China \\
  \texttt{\{xtan9,lyzhang9\}@stu.suda.edu.cn} \\
  \texttt{gdzhou@suda.edu.cn} \\}

\begin{document}
\maketitle
\begin{abstract}
It is well known that translations generated by an excellent document-level neural machine translation (NMT) model are consistent and coherent. However, existing sentence-level evaluation metrics like BLEU can hardly reflect the model's performance at the document level. To tackle this issue, we propose a \textbf{D}iscourse \textbf{Co}hesion \textbf{E}valuation \textbf{M}ethod (DCoEM) in this paper and contribute a new test suite that considers four cohesive manners (reference, conjunction, substitution, and lexical cohesion) to measure the cohesiveness of document translations. The evaluation results on recent document-level NMT systems show that our method is practical and essential in estimating translations at the document level.
\end{abstract}

\section{Introduction}

With the maturity of sentence-level neural machine translation (NMT), document-level NMT has attracted extensive interest in recent years. Compared with sentence-level NMT models, which mainly focus on intra-sentence dependencies, excellent document-level NMT models also emphasize inter-sentence consistency and coherence.
However, widely-used sentence-level evaluation metrics like BLEU~\citep{P02-1040} and Meteor~\citep{banerjee-lavie-2005-meteor} can hardly evaluate the cohesiveness of translations from the document perspective and, therefore, cannot provide a comprehensive evaluation of document translations.

In order to make up for the shortcomings of sentence-level evaluation metrics in document translation measurement, various studies on test suites and metrics have been proposed to evaluate discourse phenomena in translations,
such as the pronoun~\citep{hardmeier2012discourse,DBLP:conf/iwslt/HardmeierF10,guillou-hardmeier-2016-protest,miculicich-werlen-popescu-belis-2017-validation,bawden-etal-2018-evaluating,muller-etal-2018-large,jwalapuram-etal-2019-evaluating},
ellipsis~\citep{voita-etal-2019-good}, lexical cohesion~\citep{wong-kit-2012-extending,gong-etal-2015-document,bawden-etal-2018-evaluating,voita-etal-2019-good}, and so on~\citep{toral2018attaining,laubli2018has,rysova2019test,vojtvechova2019sao,castilho2020same}.
These studies tend to evaluate translations based on the discourse phenomenon of the source document and fill the gap in evaluating the translation from the document perspective to some extent.
However, these studies still have the following bottlenecks: (i) Most test suites are limited in data size and require a lot of manual annotations. Existing metrics are limited in cohesion type and cannot evaluate translations comprehensively. (ii) The language norms and discourse phenomena of the source and the target language may be quite different; thus, it is unsuitable to evaluate target translations based on the source document.

Facing the above challenges, in this paper, we propose a \textbf{D}iscourse \textbf{Co}hesion \textbf{E}valuation \textbf{M}ethod (DCoEM), which evaluates the cohesiveness of document translations based on target language specification.
Specifically, we build a test suite based on the target English of \textit{tst2010-tst2015} from the IWSLT 2017~\citep{cettoloEtAl:EAMT2012} evaluation campaigns.
Inspired by the manners promoting discourse cohesion summarized by~\citet{Halliday1976CohesionIE}, the test suite we construct considers reference, conjunction, substitution, and lexical cohesion simultaneously for discourse cohesion measurement.
Since our test suite construction process requires little manual involvement, it is easy to transfer the cohesiveness evaluation to other language pairs and other document-level natural language generation tasks.
\begin{table*}
  \small
\centering
\begin{tabular}{|l|l|}
\hline
Cohesive manners & Examples \\
\hline
\multirow{2}{*}{Reference} & When \underline{\textit{he}} visited the construction site last month, \underline{\textit{Mr. Jones}} talked with the union leaders  \\ & about their safety concerns. \textbf{s1}\\
\multirow{2}{*}{Conjunction} & For the whole day he climbed up the steep mountainside, almost without stopping. \underline{\textit{And}} in \\ & all this time he met no one.  \textbf{s2}\\
Word substitution & These \underline{\textit{biscuits}} are stale. Get some fresh \underline{\textit{ones}}.  \textbf{s3}\\
Lexical cohesion & Mary ate a \underline{\textit{peach}}. She likes \underline{\textit{fruit}}.  \textbf{s4}\\
Ellipsis & Smith was the first person to leave. I was the second (\textit{person}).  \textbf{s5}\\
\hline
\end{tabular}
\caption{Examples of the five cohesive manners~\citep{Halliday1976CohesionIE} that promote cohesion of a document. }
\label{tab:example}
\end{table*}

We evaluate the cohesiveness of English translations generated by recent document-level translation models on the Chinese-English translation task to investigate the practicality of our DCoEM. The results show that our method is essential in evaluating document translations as a supplement to sentence-level evaluation metrics.

\section{Preliminaries: Discourse Cohesion}

As an essential property of a document, discourse cohesion describes the local coherence. It promotes the connection between the document and its interior by coordinating the semantic relations within or between sentences. It occurs whenever ``the interpretation of some element in the discourse is dependent on that of another''~\citep{10.2307/42945277}.
\citet{Halliday1976CohesionIE} have summarized five cohesive manners that generate coherence in a document: reference, conjunction, substitution, lexical cohesion, and ellipsis.

\textbf{Reference} explains the mutual explanatory relationship between one linguistic component and another one in a document. This is usually done with pronouns. The example sentence \textbf{s1} in Table~\ref{tab:example} is a cataphora example with the pronoun ``\textit{he}'' referring to ``\textit{Mr. Jones}''.

\textbf{Conjunction} expresses relationships between things in a sentence, connecting different clauses and combining sentences. E.g., the word ``\textit{And}'' in the sentence \textbf{s2}.

\textbf{Word substitution} means using an alternative word found in the context to replace the original text unit. As illustrated in \textbf{s3}, the word ``\textit{ones}'' can be substituted with the word ``\textit{biscuits}''.

\textbf{Lexical cohesion} refers to the cohesion generated from recognizable semantic relations between words. E.g., there is a ``hyponymy'' relationship between the words ``\textit{peach}'' and ``\textit{fruit}'' in sentence \textbf{s4}, which maintains the lexical cohesion.

\textbf{Ellipsis} is the omission of one or more words from a clause that is recoverable or inferable from context. E.g., the omitted word ``\textit{person}'' in sentence \textbf{s5} can be speculated from the context. This work does not evaluate ellipsis for two reasons: (i) Ellipsis in non-pro-drop languages is less frequent than in pro-drop languages. (ii) Ellipsis is indeed a problem in languages like Chinese. However, we study the discourse cohesion of target English; therefore, we allow ellipsis to exist as a natural phenomenon of discourse cohesion.

\section{Discourse Cohesion Evaluation}

In this section, we introduce the proposed discourse cohesion evaluation method (DCoEM), which consists of two parts: test suite construction and discourse cohesion calculation.

\subsection{Test Suite Construction}

In this work, we evaluate the cohesiveness of document translations by comparing translations with the constructed test suite of target reference documents.
Inspired by the cohesion manners summarized by~\citet{Halliday1976CohesionIE}, we annotate each target document by considering four cohesive manners (reference, conjunction, substitution, and lexical cohesion) automatically to construct the test suite for document-level translation cohesion measurement.
Concretely, according to the similarity of the cohesion manners, we build three independent sets (i.e., the LexiCoh-WordSub set, the Reference set, and the Conjunction set) of each sentence as the target document annotations. In this way, the annotation process is also a process of building these sets for each sentence.

\paragraph{LexiCoh-WordSub Set.}

Considering that word substitution is a unique embodiment of lexical cohesion, we group lexical cohesion and word substitution manners into one set in this work.
Inspired by the lexical chain theory~\citep{morris-hirst-1991-lexical}, we build the LexiCoh-WordSub set in three steps: First, we employ the NLTK~\citep{bird-2006-nltk} to perform named entity recognition.
Second, we utilize the WordNet~\citep{miller-1993-wordnet} toolkit to build synonyms, hypernyms, and hyponyms for words marked with the noun and verb labels\footnote{``NN'', ``NNS'', ``NNP'', ``VB'', ``FW'', ``NNPS'' , ``VBD'', ``VBG'', ``VBN'', ``VBP'', ``VBZ''.} by the NLTK.
Third, we annotate lexical cohesion and word substitution as follows: given a selected word in the target sentence $\tau_i$, if the word or one of its synonyms, hypernyms, or hyponyms appears in the sentences $(\tau_{i-1 }, \dots, \tau_{i-5})$, then we put the word and its corresponding synonyms, hypernyms, or hyponyms into the LexiCoh-WordSub set, \textbf{lw}$_i$, of the sentence.

\paragraph{Reference Set.} In general, \textit{reference} describes the relationship between pronouns, nouns, or noun phrases that refer to the same object in the context, and the relationships between nouns or noun phrases are more like exceptional cases in lexical cohesion and word substitution. Therefore, we take them into the LexiCoh-WordSub set for consideration, and we consider the references generated by pronouns in this set.
To evaluate the discourse cohesion generated by the referencing manner, we compare the pronouns that appeared in the translation sentence with those in the target sentence to achieve it.
To this end, we directly identify all the pronouns that appeared in the sentence according to the pronoun set in Appendix and build the Reference set \textbf{r}$_i$ for the target sentence $\tau_i$.

\paragraph{Conjunction Set.} For the conjunction set, we first select the most commonly used 75 conjunctions in the target language to form a conjunction token set (see Appendix). After that, referring to the token set, we identify all the conjunctions that appeared in the target sentence $\tau_i$ to construct the Conjunction set \textbf{c}$_i$ for the sentence.

Notably, since discourse cohesion units (elements in the three sets) that refer to the same person or thing have different expressions, it is reasonable to consider the synonyms of these units for evaluation~\citep{wong-kit-2012-extending}. Therefore, we manually build a synset-like pronoun set \textbf{r}$_i'$ for each pronoun in the reference set\footnote{For example, the synset-like pronoun set of ``i'' contains ``me'', ``my'', ``mine'', and ``myself''. The complete synsets of pronouns are presented in Appendix.}.
For the Conjunction and LexiCoh-WordSub sets, we use the WordNet~\citep{miller-1993-wordnet} toolkit to build synset for each word in the sets, obtaining \textbf{c}$_i'$ and \textbf{lw}$_i'$. In this way, each sentence is attached with the three sets as mentioned above and their corresponding synsets.

According to the above instructions, we construct the test suite based on the target English sentences in \textit{tst2010-tst2015} from the IWSLT 2017 evaluation campaigns~\citep{cettoloEtAl:EAMT2012} to evaluate the cohesiveness of English document-level translations.
The detailed statistics on the English \textit{tst2010-tst2015} and the constructed test suite are listed in Table~\ref{tab:statistics}. The codes (PyTorch) and test suite will be published at \url{https://github.com/anonymous} for other researchers to use.

\begin{table}
  \small
\centering
\begin{tabular}{|l|c|c|c|c|}
\hline
 & Doc. & Sent. & Word & Word/Sent. \\
\hline
\textit{tst2010-2015} & 89 & 8549 & 167810 & 19.63  \\
\cdashline{1-5}[0.8pt/2pt]
Pronoun & 89 & 7832 & 24075 & 3.07  \\
Conjunction & 89 & 5806 & 12295 & 2.12  \\
Lexical & 89 & 8549 & 34675 & 4.06  \\
\hline
\end{tabular}
\caption{Statistics on the constructed test suite.}
\label{tab:statistics}
\end{table}

\subsection{Evaluation Method}

Given the $i$-th translation sentence with $n$ word units $\tau_i=(\tau_{i,1}, \dots, \tau_{i,n})$, we evaluate the cohesiveness of the document translation by comparing words in each translation sentence with its corresponding constructed cohesive sets.
Specifically, the score of each cohesive set is calculated as the number of correct cohesive markers in $\tau_i$ over the total number of cohesive markers in its constructed sets.
Taking the Reference set \textbf{r}$_i$ and its corresponding synset \textbf{r}$'_i$ of the sentence $\tau_i$ as an example:
\begin{gather}
    score_{i, r} = \frac{\sum{\text{j-r}(\tau_{i,k})}}{\text{len}(\bm{{\rm r}}_i)} \\
    \text{j-r}(\tau_{i,k})=\begin{cases} 1, &  \tau_{i,k} \in \bm{{\rm r}}_i \\ 0.5, &  \tau_{i,k} \in \bm{{\rm r}}'_i \\ 0, &  \text{otherwise} \end{cases}
\end{gather}
where len($\cdot$) denotes the total number of cohesive markers annotated in the constructed set and j-r($\cdot$) is a scoring function.

Then the overall discourse cohesion score for a document translation with $N$ sentences is calculated as follows:
\begin{gather}
  score_i\!=\!\frac{\sum{(\text{j-lw}(\tau_{i,k})\!+\!\text{j-r}(\tau_{i,k})\!+\!\text{j-c}(\tau_{i,k}))}}{\text{len}(\bm{{\rm lw}}_i) + \text{len}(\bm{{\rm r}}_i) + \text{len}(\bm{{\rm c}}_i)} \\
  score\!=\!\frac{\sum{score_i}}{N}
\end{gather}
where $score_i$ is the discourse cohesion score for the $i$-th sentence of the document.

\section{Experiment}

In this section, we conduct experiments on evaluating the cohesion of English translations generated by several recent document-level NMT systems on the Chinese-English document translation task.

\begin{table*}
  \small
\centering
\begin{tabular}{|l|c|cccc|}
\hline
System & \textbf{BLEU} & \textbf{DCoEM} & LC-WS & Refere. & Conjun. \\
\hline
Transformer~\citep{10.5555/3295222.3295349} & 18.22 & 52.36 & 44.59 & 63.83 & 51.83  \\
Thumt~\citep{zhang-etal-2018-improving} & 19.87 & \textbf{58.26} & 51.38 & 68.62 & \textbf{57.39}  \\
HM-GDC~\citep{tan-etal-2019-hierarchical} & \textbf{20.51} & 55.48 & 49.61 & 65.04 & 53.33  \\
CADec~\citep{voita-etal-2019-good} & 20.22 & 58.07 & \textbf{51.61} & \textbf{69.50} & 53.91  \\
BERT-nmt~\citep{zhu2020incorporating} & 19.93 & 57.34 & 51.01 & 67.91 & 54.51  \\
\hline
\end{tabular}
\caption{Results of the five systems on different evaluation metrics. ``LC-WS'', ``Refere.'', and ``Conjun.'' refer to the discourse cohesion results on the LexiCoh-WordSub, Reference, and Conjunction set, respectively.}
\label{tab:results}
\end{table*}

\subsection{Systems}

We perform discourse cohesion evaluation for the following five NMT systems:
\begin{itemize}[leftmargin=*]
  \item \textbf{Thumt}~\citep{zhang-etal-2018-improving} uses a new context encoder for document-level context modeling, and the context information is then incorporated into the original encoder and decoder of the Transformer as an extension. Besides, a large-scale sentence-level parallel corpus is used for two-step strategy training.
  \item \textbf{HM-GDC}~\citep{tan-etal-2019-hierarchical} adopts a hierarchical model to capture global document context, and it can be integrated into both the RNNSearch and Transformer frameworks. Similar to~\citep{zhang-etal-2018-improving}, this model also utilizes a two-step strategy for model training.
  \item \textbf{CADec}~\citep{voita-etal-2019-good} employs a two-pass framework, which first translates sentences through a context-agnostic model and then refines translations with context at both source and target sides.
  \item \textbf{BERT-nmt}~\citep{zhu2020incorporating} is a BERT-fused model which first uses BERT to extract representation for input sentences and then fuses the representation into the encoder-decoder of their NMT model through attention mechanisms.
  \item \textbf{Transformer}~\citep{10.5555/3295222.3295349} is known to be the state-of-the-art model in sentence-level NMT. We perform experiments on Transformer to explore the differences between context-aware and context-free systems.
\end{itemize}

\subsection{Training Data}

We use the Ted talks corpus from the IWSLT 2017~\citep{cettoloEtAl:EAMT2012} evaluation campaigns\footnote{\url{https://wit3.fbk.eu}}, containing 1,906 documents with 226K sentence-level language pairs to train the NMT systems mentioned above.
We use \textit{dev2010} (8 documents with 879 sentence pairs) for validation and \textit{tst2010-tst2015} for testing (as detailed in Table~\ref{tab:statistics}).
It should be noted that for the Thumt, HM-GDC, and CADec systems, we follow the original papers and use a two-step strategy for model training. In other words, we first use external sentence pairs\footnote{For pre-training, we use 2.8M sentence pairs from corpora LDC2003E14, LDC2004T07, LDC2005T06, LDC2005T10 and LDC2004T08 (Hongkong Hansards/Laws/News).} to pre-train the NMT models and then use the document-level corpus to fine-tune the models.

\subsection{Results}

We evaluate translations generated by the context-aware models and the Transformer model mentioned above with the BLEU (multi-bleu.perl) and the proposed DCoEM, as shown in Table~\ref{tab:results}.

From the results, all the context-aware models greatly surpass the Transformer on BLEU and DCoEM, demonstrating the capability of context-aware models in generating document translations.
However, by observing the performance of sentence- and document-level metrics obtained by the context-aware models, we find that the score of discourse cohesion evaluation (DCoEM) is not proportional to that of sentence evaluation (BLEU). In other words, a high sentence evaluation score for a document translation is not equivalent to a high discourse cohesive evaluation score (lines 2 and 3). This indicates that using BLEU to estimate the document-level NMT model is not enough, and our DCoEM is necessary as a supplement for the sentence-level metric.
To verify the reliability of our DCoEM, we also perform human evaluation, and all the accuracy scores are higher than 88\%.
Moreover, although existing context-aware models can make up for the shortcomings of context-free models in document-level translation, the overall results indicate that the current document-level NMT models still have a long way to go to improve the cohesion of translations.

\section{Conclusion}

In this work, we built a test suite with various cohesion manners considered and proposed a discourse cohesion evaluation method (DCoEM) to evaluate document translations with English as the target language.
Experiments on recent document-level NMT systems show that discourse cohesion evaluation, like our method, is necessary to make up for the shortcomings of sentence-level metrics in evaluating document translations.


\bibliography{anthology,custom}

%
\section{Appendix}
\subsection*{A. System Settings}
In this work, we replicated five previous state-of-the-art NMT systems and manually kept the model configurations (e.g., number of parameters in each model, batch size, training and evaluation runs, etc.) the same as theirs for a fair comparison. Notably, all the replication experiments were conducted on the NVIDIA Tesla P40 GPUs.

\subsection*{B. Conjunction}

\begin{table}[h]
  \small
\centering
\begin{tabular}{|l|}
\hline
Conjunctions \\
\hline
and, also, too, furthermore, moreover, again, another, \\ third, fourth, now, then, before, after, immediately, soon, \\ next, gradually, suddenly, nor, far, behind, beyond, above, \\ below, around, outside, still, however, since, otherwise, \\ indeed, surely, necessarily, certainly, truly, similarly, \\ likewise, as, besides, even, first, initially, second, earlier, \\ later, following, afterwards, thus, therefore, consequently, \\ thereby, eventually, nonetheless, obviously, plainly, \\ undoubtedly, if, unless, whether, until, meanwhile, lastly, \\ finally, but, for, so, because, nor, yet \\
\hline
\end{tabular}
\caption{Conjunctions used in the construction of our test suite.}
\label{tab:accents}
\end{table}

\subsection*{C. Pronoun}

\begin{table}[hp]
  \small
\centering
\begin{tabular}{|ll|}
\hline
Pronouns & Synsets\\
\hline
i & me, my, mine, myself \\
me & i, my, mine, myself \\
my & i, me, mine, myself \\
mine & i, me, my, myself \\
we & us, our, ours, ourselves \\
us & we, our, ours, ourselves \\
our & us, we, ours, ourselves \\
ours & us, we, our, ourselves \\
you & your, yours, yourself, yourselves \\
your & you, yours, yourself, yourselves \\
yours & you, your, yourself, yourselves \\
he & him, his, himself \\
him & he, his, himself \\
his & he, him, himself \\
she & her, hers, herself \\
her & she, hers, herself \\
it & its, itself \\
its & it, itself \\
they & them, their, theirs, themselves \\
them & they, their, theirs, themselves \\
their & them, they, theirs, themselves \\
theirs & them, they, their, themselves \\
one & ones \\
ones & one \\
myself & i, me, my, mine \\
yourself & you, your, yours, yourselves \\
yourselves & you, your, yours, yourself \\
himself & him, his, her \\
herself & she, her, hers \\
itself & it, its \\
ourselves & we, us, our, ours \\
themselves & them, they, their, theirs \\
this & that, it, one \\
that & this, it, one \\
these & those \\
those & these \\
who & whom, whose, whoever, whomever, whosever \\
whom & who, whose, whoever, whomever, whosever \\
whose & who, whom, whoever, whomever, whosever \\
what & whatever \\
which & whichever \\
whichever & which \\
whatever & what \\
whoever & who, whom, whomever, whosever, whose \\
whosever & who, whom, whomever, whoever, whose \\
whomever & whoever, who, whom, whose, whosever \\
some & many, any \\
many & some, any \\
any & some, many \\
both & ---\\
\hline
\end{tabular}
\caption{Pronouns and its synsets used in the construction of our test suite.}
\label{tab:accents}
\end{table}
%

\end{document}